%% file: main.tex
\documentclass[runningheads]{llncs}

\usepackage{eccv}

\usepackage{eccvabbrv}

\usepackage{graphicx}
\usepackage{booktabs}

\usepackage[accsupp]{axessibility}  

\usepackage{hyperref}

\usepackage{orcidlink}

\usepackage{multirow}
\usepackage{xcolor}

\begin{document}

\title{Setting the Stage: Text-Driven Scene-Consistent Image Generation} 

\titlerunning{Setting the Stage: Text-Driven Scene-Consistent Image Generation}

\author{
Cong Xie\thanks{Equal contribution.} \and
Che Wang\protect\footnotemark[1] \and
Yan Zhang \and
Ruiqi Yu \and
Han Zou \and
Zheng Pan \and
Zhenpeng Zhan
}

\authorrunning{C.~Author et al.}

\institute{Global Business Unit, Baidu Inc.}

\maketitle

\begin{figure}
  \centering
  \includegraphics[width=1\textwidth]{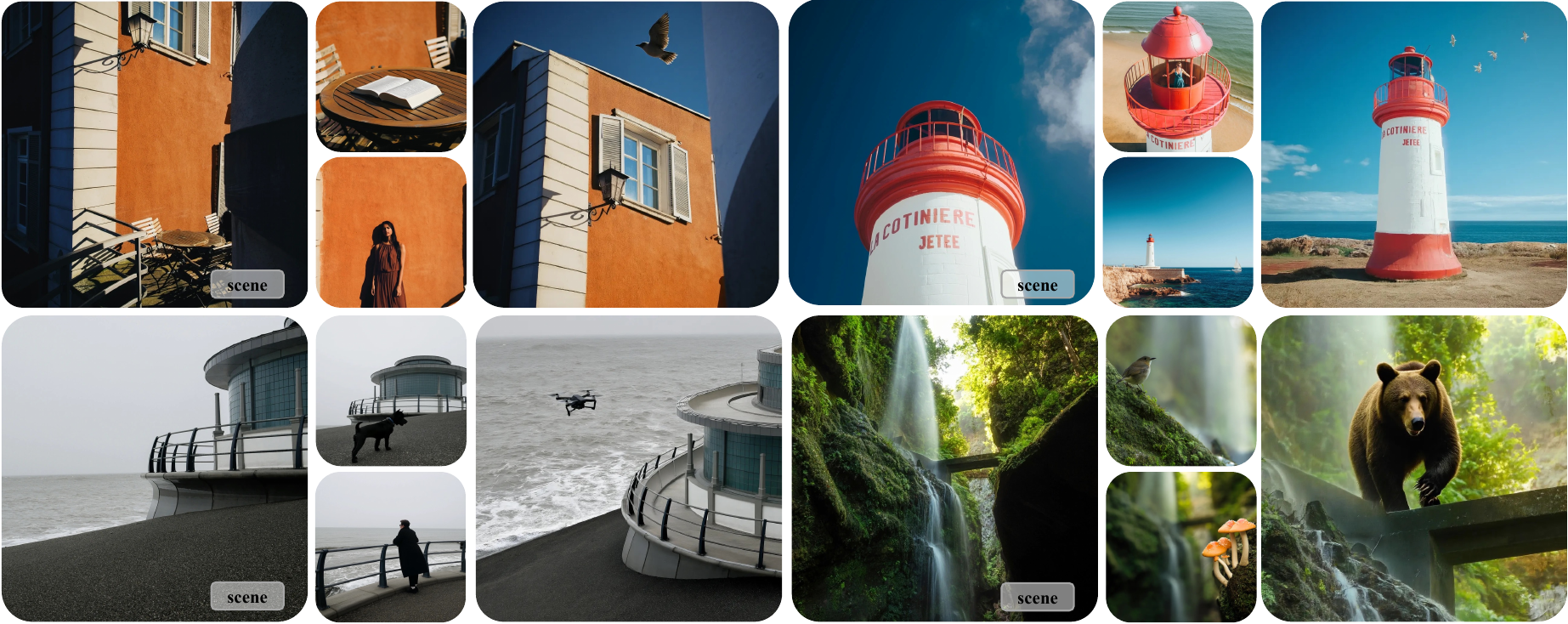}
  \caption{Scene Staging: Text-driven scene-consistent image generation. Images labeled with the subscript "scene" are reference scene images used to guide generation; the remaining images are our results, which depict diverse viewpoints and compositions while faithfully following the textual instructions and preserving the reference scene identity, rather than merely copying the reference.}
   \label{fig:main}
\end{figure}

\input{sec/0_abstract}
\input{sec/1_introduction}
\input{sec/2_relatedwork}
\input{sec/3_methods}
\input{sec/4_experiments}
\input{sec/5_conclusion}
\appendix
\input{sec/X_suppl}

%
%
\bibliographystyle{splncs04}
\bibliography{main}
\end{document}

%% file: sec/0_abstract.tex
\begin{abstract}
  We focus on the foundational task of Scene Staging: given a reference scene image and a text condition specifying an actor category to be generated in the scene and its spatial relation to the scene, the goal is to synthesize an output image that preserves the same scene identity as the reference image while correctly generating the actor according to the spatial relation described in the text. 
  Existing methods struggle with this task, largely due to the scarcity of high-quality paired data and unconstrained generation objectives. 
  To overcome the data bottleneck, we propose a novel data construction pipeline that combines real-world photographs, entity removal, and image-to-video diffusion models to generate training pairs with diverse scenes, viewpoints and correct entity-scene relationships. 
      Furthermore, we introduce a novel correspondence-guided attention loss that leverages cross-view cues to enforce spatial alignment with the reference scene. Experiments on our scene-consistent benchmark show that our approach achieves better scene alignment and text–image alignment than state-of-the-art baselines, according to both automatic metrics and human preference studies. Our method generates images with diverse viewpoints and compositions while faithfully following the textual instructions and preserving the reference scene identity.
  \keywords{Scene Consistency \and Text to Image Generation  \and Image Editing}
\end{abstract}

%% file: sec/1_introduction.tex
\section{Introduction}
\label{sec:Introduction}

The demand for visual storytelling, such as comic generation and video storyboarding, has surged in recent years. 
In the production of visual stories, the creation of narrative imagery can be viewed as a two-step process: first, setting the stage (the background scene), and second, placing the actors (character or object).
Existing image editing models~\cite{wu2025qwen,seedream2025seedream,li2025uniworld,xia2025dreamomni2} have made significant progress in local image editing, such as inserting or replacing entities while largely preserving the overall image composition. Therefore, they are well suited for the second step: given an already established stage image that does not require major compositional changes, the reference actor can be inserted or replaced accordingly.

In this work, we focus on the stage setting step: generating an image conditioned on a reference scene image and textual instructions that specify the actor category (e.g., a man or a book) and its spatial relationship within the scene.
In visual storytelling, scenes often need to be depicted from different viewpoints or compositions with respect to the actor, which introduces two challenges:
(1) instruction following: the generated image should adapt the reference scene to the desired viewpoint and composition specified by the text rather than simply replicate it;
(2) scene consistency: the generated images should preserve the same scene identity as the reference to maintain narrative coherence. 

Existing methods~\cite{yang2023paint,blackforest2024fluxtools,ju2024brushnet,wu2025qwen,seedream2025seedream}, struggle to balance scene consistency with instruction following, either replicating the reference scene with high consistency but poor responsiveness to the prompt, or prioritizing prompt compliance at the expense of scene consistency. Such limitations largely stem from both data availability and training objectives.
From a data perspective, effective training pairs for this task should satisfy \textbf{three key properties}:
(1) the input and target images share the same scene identity but differ in viewpoint and composition; (e.g., the text may request a top-down view of the scene even when the reference image is a low-angle view, as in the lighthouse example in Figure~\ref{fig:main})
(2) the actor is absent from the input image but correctly placed in the target image;
(3) the training data contains a large and diverse set of scenes.

It is non-trivial to construct a training data fulfilling all the three properties described above. 
Existing multi-view scene datasets, such as DL3DV-10K~\cite{ling2024dl3dv} , only satisfy Property (1). One may attempt to place actors into the target-view image using inpainting or image editing models. However, the generated results often suffer from scale mismatch, lighting inconsistency, and blending artifacts along object boundaries. Such artifacts break the correct entity–scene relationships and degrade the quality of supervision.
Another option is to mine frame pairs from web videos. However, shot changes and scene transitions frequently result in frame pairs that do not depict the same scene at all. Restricting the sampling to adjacent frames can preserve scene identity, but it significantly reduces the viewpoint baseline, yielding pairs with nearly identical spatial layouts. As a result, such data fails to teach the model how to maintain scene consistency under meaningful viewpoint or compositional changes.

To overcome the limitations of existing datasets, we introduce a scene-consistent data construction pipeline designed to fulfill all three properties. 
Specifically, we directly use real-world photographs as training targets to satisfy Properties 2 and 3. We apply Entity Removal to obtain clean background images while preserving realistic scene layouts, further supporting Property 2. Finally, we leverage video generation models to synthesize spatially diverse yet scene-consistent views, which enables us to satisfy Property 1.
Building on these data, we further propose a novel correspondence-guided attention loss to explicitly optimize the network for scene preservation. To address the unconstrained nature of standard diffusion models during cross-view generation, we leverage cross-view correspondence cues to directly supervise the intermediate joint-attention layers of the diffusion backbone. By regularizing the attention score matrix, this loss forces tokens that correspond to the same scene location across different views to allocate higher attention weights to each other. Consequently, by encouraging the joint attention to place high mass on corresponding scene regions, this explicit supervision compels the model to preserve the underlying scene structure and appearance, even as the viewpoint and composition undergo significant text-driven modifications.

Our contributions are summarized as follows:
\begin{itemize}
    \item We propose a novel and scalable scene-consistent data generation pipeline. This pipeline overcomes the key obstacles in constructing high-quality training pairs that guarantee large scene diversity, correct entity-scene relationships, and rich viewpoint variations.
    \item We introduce a novel correspondence-guided attention loss that regularizes the attention score matrix, forcing tokens corresponding to the same scene location across different views to allocate higher attention weights to each other. This mechanism guarantees robust scene preservation even when the narrative introduces significant text-driven viewpoint and composition shifts.
    \item We demonstrate that our method consistently outperforms existing baselines, achieving a superior balance between accurate text instruction following and scene consistency. Extensive experiments show that the generated images faithfully reflect the textual instructions while preserving the given reference scene identity. This advantage is validated through both automatic quantitative metrics and comprehensive human evaluations.
\end{itemize}

%% file: sec/2_relatedwork.tex
\section{Related Works}
\label{sec:relatedwork}

\noindent\textbf{Text-Driven Image Editing.}
Existing image editing models~\cite{seedream2025seedream,li2025uniworld,xia2025dreamomni2} are typically designed as multi-task, instruction-following general frameworks. For example, FLUX.1 Kontext~\cite{labs2025flux} unifies image generation and editing within a single flow matching model via sequence concatenation in the latent space. ACE++~\cite{mao2025ace++} introduces an instruction-based diffusion framework that handles diverse editing tasks through context-aware content filling. However, these versatile frameworks lack explicit training for scene-oriented tasks. Consequently, they frequently encounter a severe trade-off between environment preservation and instruction adherence: they tend to either rigidly reconstruct the original scene while failing to reflect the text prompt, or aggressively execute semantic modifications at the expense of the underlying scene's structural consistency.

\noindent\textbf{Image Inpainting.} 
Local image inpainting\cite{yang2023paint,blackforest2024fluxtools} primarily targets filling missing regions or adding local content to an existing image. While early deep learning approaches like MAT~\cite{li2022mat} utilized transformers for large hole filling, they specifically customized the attention mechanism to aggregate non-local information only from valid tokens, enabling efficient high-resolution processing. With the advent of diffusion models, text-guided inpainting introduced much richer semantic controllability. For instance, SmartBrush~\cite{xie2023smartbrush} incorporates both text and shape guidance, predicting object masks during synthesis to explicitly preserve the surrounding background. To further enhance generation quality and applicability, BrushNet~\cite{ju2024brushnet} proposes a plug-and-play, decomposed dual-branch diffusion architecture. By explicitly separating masked image feature extraction from the noisy latent generation, it achieves dense per-pixel constraints and maintains strong structural fidelity. However, these methods inherently rely on fixed spatial layouts. Consequently, when a narrative demands global shifts in perspective or composition, they cannot reliably transform the overall scene while preserving the underlying environment.

\noindent\textbf{Consistent Image Generation.}
Consistent image generation has witnessed rapid development in recent years. A multitude of methods, ranging from adapter-based conditional models (e.g., IP-Adapter~\cite{ye2023ip}, InstantID~\cite{wang2024instantid}, PhotoMaker~\cite{li2024photomaker}) and attention-tuning mechanisms (e.g., StoryDiffusion~\cite{zhou2024storydiffusion}, StoryMaker~\cite{zhou2024storymaker}) to efficient autoregressive frameworks (e.g., Infinite-Story~\cite{park2025infinite}), have been proposed to maintain fine-grained character attributes and global styles across multiple generated frames. 
While extensive research has successfully addressed character (actor) consistency, scene (stage) consistency remains a significant bottleneck. SceneDecorator~\cite{song2025scenedecorator} emerges as the most relevant baseline, employing a training-free global-to-local planning strategy. However, such training-free attention sharing relies heavily on raw visual feature similarity. It degrades significantly when the narrative introduces substantial viewpoint or layout changes, as the un-tuned attention mechanisms fail to align drastically altered visual features. In contrast, our method focuses on the foundational task of Scene Staging: seamlessly embedding a text-specified entity into a reference scene under dynamic viewpoint and composition changes, while accurately reflecting spatial relationships and preserving the original structural integrity of the environment.

%% file: sec/3_methods.tex
\section{Methods}
\label{sec:method}

\subsection{Overview}

\begin{figure}[t]
  \centering
  \includegraphics[width=1\textwidth]{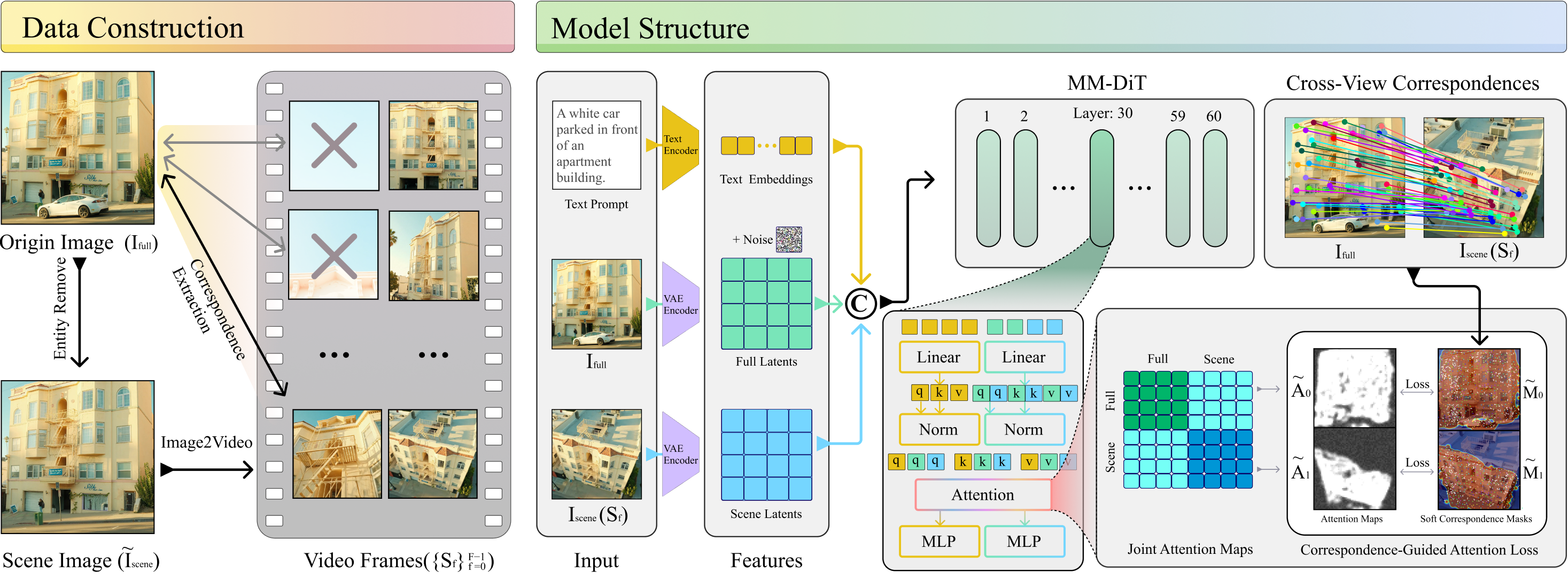}
  \caption{Overall framework of our method with two contributions: (a) Scene-Consistent Data Construction Pipeline, generating high-quality, scene-consistent paired data with large scene diversity, correct entity-scene relationships, and rich viewpoint variations; (b) Correspondence-Guided Attention Loss, encouraging attention to focus on spatially corresponding regions to better preserve scene structure.}
   \label{fig:overview}
\end{figure}

Given a reference scene image $I_{\mathrm{scene}}$ and a text condition $c$ that specifies both the actor category (e.g., a man or a book) to be placed in the scene and its spatial relationship to the scene, our goal is to generate a scene-consistent output image $I_{\mathrm{out}}$ that (i) depicts the same scene identity as $I_{\mathrm{scene}}$, and (ii) correctly generates the actor according to the spatial relation described in the text.
We denote the generative model as
\begin{equation}
    I_{\mathrm{out}} = f_\theta(I_{\mathrm{scene}}, c),
\end{equation}
where $f_\theta$ is implemented as a diffusion-based image generator conditioned on both the reference scene and the text.

As illustrated in Figure~\ref{fig:overview}, our method has two components. First, we build a scene-consistent data construction pipeline that converts real world images into multi-view tuples comprising entity-removed scenes, synthesized scene views, and soft correspondence masks. Second, during training we apply a correspondence-guided attention loss that uses these correspondence cues to regularize intermediate attention maps, encouraging attention to focus on spatially corresponding regions across views and thereby better preserving scene structure.

\subsection{Scene-Consistent Data Construction}
\label{subsec:data_synthesis}
As described in the Introduction, ideal training data should fulfill three properties. Next, we introduce a data construction pipeline that satisfies all these properties.

\noindent\textbf{Real-world Photographs as Supervision to Fulfill Properties 2 and 3}.
We start by collecting real-world photographs containing humans and generic objects in diverse scenes. These images $I_{\mathrm{full}}$ do not need to be paired multi-view observations; single-view photographs are sufficient. Such data is easy to collect at scale, enabling a large-scale and diverse set of scenes to satisfy Property 3. Moreover, since these are real photographs, the objects are naturally and correctly placed in the scene, which partially satisfies Property 2.

\noindent\textbf{Entity Removal to Fulfill Property 2.}
For each image $I_{\mathrm{full}}$, we automatically erase a target entity region (person or object) to obtain an entity-removed scene $\tilde{I}_{\mathrm{scene}}$.
For human-centric photos, we apply FLUX.1 Kontext Dev~\cite{labs2025flux} for person removal, and additionally apply face detection~\cite{guo2021sample} and human segmentation~\cite{liu2024grounding, ravi2024sam} to discard low-quality results and maintain high-quality entity-removed scenes.
For non-human objects, we use open-source image-editing datasets OmniEdit~\cite{wei2024omniedit} that provide object-removal image pairs.
The resulting $\tilde{I}_{\mathrm{scene}}$ images serve as realistic scene-only inputs for the subsequent multi-view synthesis stage.

\noindent\textbf{Multi-View Scene Generation via Image-to-Video to Fulfill Property 1.}
Given $\tilde{I}_{\mathrm{scene}}$, we employ a pre-trained image-to-video diffusion model $\mathcal{G}_{\mathrm{i2v}}$ (Wan 2.2~\cite{wan2025wan}) to synthesize a short video sequence
\begin{equation}
    \{ S_f \}_{f=0}^{F-1} = \mathcal{G}_{\mathrm{i2v}}(\tilde{I}_{\mathrm{scene}}),
\end{equation}
where $F$ is the total number of frames, and each frame $S_f$ corresponds to the same underlying scene under a slightly different camera motion.

Benefiting from large-scale training on real-world videos, the image-to-video diffusion model $\mathcal{G}_{\mathrm{i2v}}$ can transform a single high-quality photograph into a short clip with spatial coherence and smooth camera motion. In our pipeline, this allows any still image to be converted into a compact multi-view sequence, greatly expanding the pool of training scenes and naturally enriching scene viewpoints.

Instead of pairing with all synthesized frames, we first filter out those exhibiting severe video hallucinations or structural drift. By evaluating the robust inliers obtained from the feature matching network~\cite{ren2025minima}, we discard low-quality outputs and pair the source image $I_{\mathrm{full}}$ only with one of the surviving high-quality frames $S_f$ to form a scene pair $(I_{\mathrm{full}}, S_f)$.
In our training formulation, $S_f$ plays the role of $I_{\mathrm{scene}}$, while $I_{\mathrm{full}}$ serves as a proxy for $I_{\mathrm{out}}$.

\noindent\textbf{Cross-View Correspondence Extraction.}
To obtain explicit scene-level correspondence cues between $I_{\mathrm{full}}$ and the selected scene view $S_f$, we apply a pre-trained feature matching network~\cite{ren2025minima} that estimates sparse correspondences between two views of the same scene under viewpoint changes.
Given the pair $(I_{\mathrm{full}}, S_f)$, the matcher outputs
\begin{equation}
    \{ (x_i^0, y_i^0) \}_{i=1}^{N}, \quad
    \{ (x_i^1, y_i^1) \}_{i=1}^{N},
\end{equation}
where $(x_i^0, y_i^0)$ and $(x_i^1, y_i^1)$ denote coordinates of matched points in $I_{\mathrm{full}}$ and $S_f$, respectively.

We convert these sparse matches into dense, soft correspondence masks.
Let $H \times W$ be the image resolution and $K \in \mathbb{R}^{(2r+1)\times(2r+1)}$ be a radial kernel (e.g., Gaussian) whose values decay with distance from the center.
For each match $((x_i^0,y_i^0),(x_i^1,y_i^1))$, we add a copy of $K$ centered at $(x_i^0,y_i^0)$ and $(x_i^1,y_i^1)$ onto masks $M_0, M_1 \in \mathbb{R}^{H \times W}$, respectively, accumulating overlapping contributions and clipping them to a fixed range.
We then crop and downsample $M_0$ and $M_1$ to the spatial resolution of the latent feature maps used for attention supervision, obtaining
\begin{equation}
    \tilde{M}_0, \tilde{M}_1 \in \mathbb{R}^{H_s \times W_s}.
\end{equation}
High responses in $\tilde{M}_0$ and $\tilde{M}_1$ highlight regions that are spatially corresponding between the two views.

\subsection{Correspondence-Guided Attention Loss}
\label{subsec:geometry_aware}

We build on the state-of-the-art image editing model Qwen-Image-Edit~\cite{wu2025qwen} and augment it with correspondence-guided attention supervision.

\noindent\textbf{MMDiT Backbone.}
Our model adopts the double-stream Multimodal Diffusion Transformer (MMDiT) architecture of Qwen-Image-Edit, jointly modeling noise and image latents under text conditioning. Given a noisy latent of the target image $I_{\mathrm{full}}$ (proxy for $I_{\mathrm{out}}$), a reference scene image $I_{\mathrm{scene}}$, and a text prompt $c$ describing the inserted entity, the model predicts the added noise by using two streams: one for image tokens, and another for conditioning tokens (which jointly encode the scene and the text).

We keep the variational autoencoder (VAE), text encoder, and backbone weights of Qwen Image Edit frozen, and only introduce a small number of trainable parameters via lightweight adapters. This parameter-efficient fine-tuning preserves the strong priors of the base model while enabling specialization to our scene staging objective and the proposed correspondence-guided attention regularization.

\noindent\textbf{Correspondence-Guided Attention Loss.}
Given the source image $I_{\mathrm{full}}$ (denoted as view 0) and the
scene condition $S_f$ (denoted as view 1), we extract their visual
tokens from an intermediate attention layer of MMDiT,
$
z_0 \in \mathbb{R}^{L_0 \times d}$, $z_1 \in \mathbb{R}^{L_1 \times d}$, where $L_0$ and $L_1$ denote the number of tokens for view 0 and view 1,
respectively, and $d$ is the token feature dimension.
We then form a joint token sequence
$X = [z_0; z_1] \in \mathbb{R}^{(L_0 + L_1) \times d}$,
and use it as both queries and keys, $Q = K = X$.
The attention score matrix is computed as
\begin{equation}
A = \operatorname{softmax}\!\left(\frac{QK^\top}{\sqrt{d}}\right)
\in \mathbb{R}^{(L_0 + L_1) \times (L_0 + L_1)}.
\end{equation}

We denote by $\mathcal{I}_0 = \{1,\dots,L_0\}$ the index set of tokens
from view~0 and by $\mathcal{I}_1 = \{L_0+1,\dots,L_0+L_1\}$ the index
set of tokens from view~1. Each entry $A_{ij}$ represents the attention
weight assigned by token $i$ to token $j$. In particular, for
$i \in \mathcal{I}_0$ and $j \in \mathcal{I}_1$, $A_{ij}$ encodes the
attention from a token in view~0 to a token in view~1, while
$A_{ji}$ with $j \in \mathcal{I}_1$ and $i \in \mathcal{I}_0$ encodes
the reverse direction.
In our setting, we expect tokens that correspond
to the same scene location across views to allocate higher attention to
each other, so that spatially matching regions are emphasized in the
cross-view attention map. 
If a token $p \in \mathcal{I}_0$ participates in a successful keypoint correspondence, its counterpart in $\mathcal{I}_1$ will exhibit a similar visual feature. As a result, both the attention that $p$ allocates to tokens $q \in \mathcal{I}1$ and the attention it receives from them tend to be higher. We therefore summarize its cross-view interaction by symmetrizing and averaging these two directions as

\begin{equation}
a^{(0)}_p
= \frac{1}{2}\bigl(\frac{1}{L_1} \sum_{q \in \mathcal{I}_1}A_{p q} + \frac{1}{L_1} \sum_{q \in \mathcal{I}_1}A_{q p}\bigr),
\quad
q \in \mathcal{I}_1.
\end{equation} 

Since the keypoint-based correspondence map $\tilde{M}_0$ assigns a high value to the location of token $p$, we use it as supervision in our loss, encouraging $a^{(0)}_p$ to take a similarly high value:

\begin{equation}
\mathcal{L}_{\text{attn}0}
=
\frac{1}{N}
\left(
\left\| \tilde{A}_0 - \tilde{M}_0 \right\|_2^2
\right),\quad
(\tilde{A}_0)_p = a^{(0)}_p,
\label{eq:attn-loss1}
\end{equation}
where $N = H_s W_s$ denotes the number of spatial locations per view.

Similarly, we apply the same supervision in the reverse direction from $q$ to $p$, leading to the following attention loss:
\begin{equation}
\mathcal{L}_{\text{attn}1}
=
\frac{1}{N}
\left(
\left\| \tilde{A}_1 - \tilde{M}_1 \right\|_2^2
\right).
\label{eq:attn-loss2}
\end{equation}

The correspondence-guided attention loss is defined as
\begin{equation}
\mathcal{L}_{\text{attn}}
=
\mathcal{L}_{\text{attn0}}
+
\mathcal{L}_{\text{attn1}}
.
\label{eq:attn-loss}
\end{equation}

This loss encourages the joint attention to place high mass on spatially corresponding scene regions in both views, guiding the model to preserve underlying scene structure while editing according to the text and to produce viewpoint-consistent content.
In practice, we apply this supervision only to a single mid-level joint-attention block, which empirically balances strong spatial guidance with sufficient flexibility in the remaining layers.

\subsection{Training Objective}
\label{subsec:objective}
We train the model using a combination of the flow matching loss and the correspondence-guided attention loss. We construct the intermediate noisy latent of the target image $I_{\mathrm{full}}$ at a randomly sampled timestep $t \in [0, 1]$, and train the network to predict the corresponding vector field given the noisy latent, the reference scene $I_{\mathrm{scene}}$, and the text $c$.
Let $u_0$ denote the clean latent representation of $I_{\mathrm{full}}$, and $u_1 \sim \mathcal{N}(0, I)$ denote the initial Gaussian noise. The intermediate noisy state $u_t$ is constructed via linear interpolation: $u_t = t u_0 + (1 - t) u_1$. The model prediction $v_\theta$ is trained to match the ground-truth velocity $v_t = u_0 - u_1$.
The flow matching loss is formulated as:
\begin{equation}
\mathcal{L}_{\mathrm{flow}} = \mathbb{E}_{t, u_0, u_1} \left\| v_\theta(u_t, t, I_{\mathrm{scene}}, c) - v_t \right\|^2. 
\end{equation}

The total training objective is
\begin{equation}
    \mathcal{L} =
    \mathcal{L}_{\mathrm{flow}}
    +
    \lambda \, \mathcal{L}_{\mathrm{attn}},
\end{equation}
where $\lambda$ controls the strength of the correspondence-guided attention supervision.
In our experiments, we use a fixed $\lambda$ (e.g., $\lambda = 3.0$), and apply $\mathcal{L}_{\mathrm{attn}}$ only to selected attention blocks.

%% file: sec/4_experiments.tex
\section{Experiments}

\subsection{Implementation Details}
The training set is constructed from Places365~\cite{zhou2017places} and OmniEdit~\cite{wei2024omniedit}, resulting in
96{,}040 distinct entity-centric scene images.
For each image, we synthesize a short video clip using Wan~2.2~\cite{wan2025wan} and filter the frames using MINIMA~\cite{ren2025minima}, a RoMa-based~\cite{edstedt2024roma} dense matcher, by enforcing over $2000$ robust inliers to mitigate video hallucination and structural drift.
On average, each source image contributes about 14 frames,
yielding 1{,}336{,}509 scene pairs in total. Text descriptions for all samples are annotated using the
GLM-4.1V-Thinking vision–language model~\cite{hong2025glm}.
Crucially, during data loading, to prevent the model from learning trivial identity mappings, we employ a temporally-weighted sampling strategy. Based on the prior that later frames inherently exhibit larger cumulative camera motions, they are dynamically assigned higher selection probabilities within each video clip.

Training is conducted on 8 NVIDIA H100 GPUs with a per-GPU batch size of 1 and gradient accumulation over 6 steps, resulting in an effective batch size of 48.
We adopt multi-resolution training
with arbitrary aspect ratios, constraining the longer image side to at most 1024 pixels
(maximum resolution $1024 \times 1024$). Optimization proceeds for 4{,}000 steps. At inference
time, we use 28 denoising steps.

\subsection{Evaluation Setup}
\noindent\textbf{Benchmark.}
The test set comprises 425 images collected from a variety of scene types. 
For each image, five distinct textual prompts are created using Gemini-2.5-Pro~\cite{comanici2025gemini}, yielding 
a total of 2{,}125 scene-prompt pairs. 
All evaluated models are tested on this identical set of inputs to ensure comparability under consistent conditions.

\begin{figure}[htbp]
  \centering
  
  \includegraphics[width=1\textwidth]{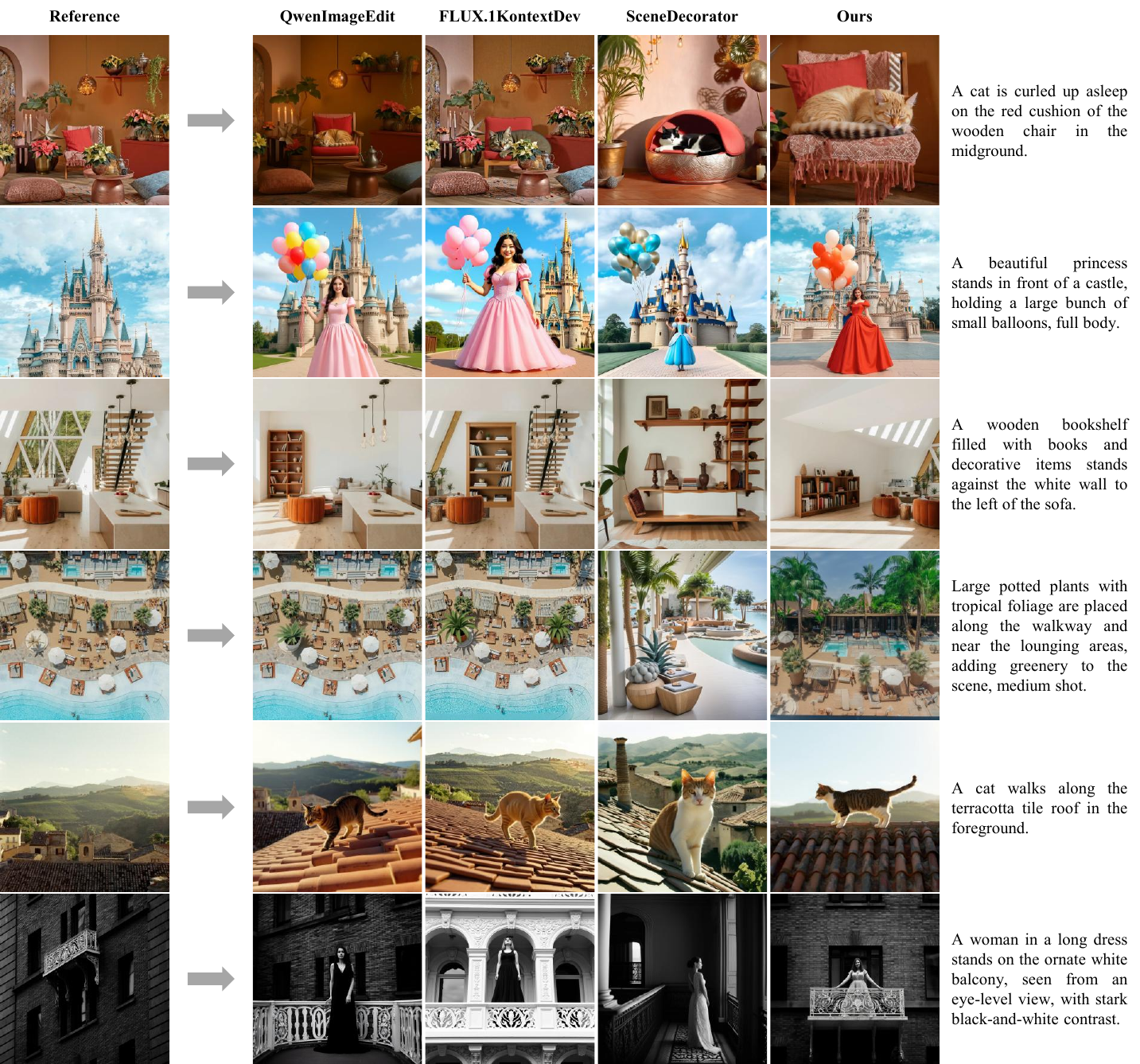}
  \caption{Qualitative comparison of our method with other baselines. Our method demonstrates superior scene consistency and text-image consistency compared to other baselines.}
  \label{fig:qualitive_compare}
  
  \vspace{1.5em} 
  
  \captionof{table}{Quantitative comparison of automatic metrics and user study across baselines. The best result is highlighted in \textbf{bold}.}
  \label{tab:QuantitiveCompare}
  \begin{tabular}{@{}lcccc@{}}
  \toprule
  \multirow{2}{*}{Methods} & \multicolumn{2}{c}{Automatic Metrics} & \multicolumn{2}{c}{User Study} \\ 
  \cmidrule{2-3} \cmidrule{4-5} 
  & G2.5F-SA$\uparrow$ & G2.5F-TIA$\uparrow$ & Scene Align$\uparrow$ & Text Align$\uparrow$ \\
  \midrule
  Qwen-Image-Edit~\cite{wu2025qwen} & 9.434 & 9.306 & 34.89\% & 28.43\% \\
  FLUX.1 Kontext Dev~\cite{labs2025flux} & 8.395 & 8.841 & 26.29\% & 25.46\% \\
  SceneDecorator~\cite{song2025scenedecorator}     & 4.104 & 8.176 & 1.69\%  & 17.67\% \\
  Ours               & \textbf{9.811} & \textbf{9.639} & \textbf{37.13\%} & \textbf{28.45\%} \\
  \bottomrule
  \end{tabular}
\end{figure}

\noindent\textbf{Evaluation Metrics.}
We assess the quality of the generated image using Gemini 2.5 Flash~\cite{comanici2025gemini}, focusing on two aspects: 
(i) Gemini 2.5 Flash scene alignment (G2.5F-SA),  which measures the scene consistency between the reference image and the generated frame. It evaluates whether the model preserves the spatial structural layout of the scene while ignoring acceptable modifications.
Variations such as object modifications, depth-of-field effects, zoom or rotation, and moderate viewpoint changes are considered acceptable modifications and are excluded from negative judgment, with the score focusing solely on scene background consistency.
(ii) Gemini 2.5 Flash text-image alignment (G2.5F-TIA), evaluates how faithfully the visuals follow the given textual prompt.
Both metrics are scored on a scale from 0 (poor) to 10 (excellent).
Previous work mainly relies on CLIP-T~\cite{gal2022image} and DreamSim~\cite{fu2023dreamsim} for evaluation, but we find that these metrics do not accurately capture our task; a detailed analysis and the specific evaluation prompts used for Gemini 2.5 Flash are provided in the supplementary material.

\subsection{Comparison with State-of-the-Art Methods}

\noindent\textbf{Quantitative Comparisons.}
We conduct a comparative evaluation of our method with three representative baselines: 
Qwen-Image-Edit~\cite{wu2025qwen}, FLUX.1 Kontext Dev~\cite{labs2025flux}, and SceneDecorator~\cite{song2025scenedecorator}.
The performance is measured using objective automatic metrics, with detailed results summarized in Table~\ref{tab:QuantitiveCompare}.
As shown in the table, our method achieves the highest performance across all automatic metrics. While general-purpose editing models like Qwen-Image-Edit and FLUX.1 Kontext Dev demonstrate strong foundational capabilities, they yield lower scores in both scene alignment and text-image alignment compared to our method. We attribute these performance gaps primarily to their reliance on generic training objectives and the absence of fine-tuning on datasets explicitly curated for scene consistency. More remarkably, SceneDecorator suffers a degradation in scene alignment, which empirically validates our earlier hypothesis: training-free attention sharing relies heavily on raw visual similarity, causing the underlying structural integrity to collapse when dynamic camera movements alter the visual features. In contrast, by explicitly regularizing spatial alignment through correspondence guidance, our method successfully maintains the background stage while accurately executing the prompt.

\noindent\textbf{Qualitative Comparisons.}
The visualization results are presented in Figure~\ref{fig:qualitive_compare}, with additional examples available in the supplementary material.
While Qwen-Image-Edit~\cite{wu2025qwen} and FLUX.1 Kontext Dev~\cite{labs2025flux} effectively follow textual instructions, they often suffer from "contextual hallucination"---introducing unintended changes to background textures, scene layouts, or other elements during entity insertion---and demonstrate limited spatial alignment capability.
SceneDecorator~\cite{song2025scenedecorator}, while maintaining a coherent global atmosphere, often exhibits degradation in fine-grained visual details. As observed in  Figure~\ref{fig:qualitive_compare}, specific background elements (e.g., material textures, wall patterns, and intricate objects) tend to lose fidelity or undergo semantic drift compared to the reference scene. This leads to a noticeable divergence from the original scene identity, particularly in complex environments.
In contrast, our method seamlessly integrates new entities while preserving the fine-grained textural richness and structural integrity of the reference "stage".

\noindent\textbf{User Study.}
We carried out a user study on the 2{,}125 scene–prompt pairs using the corresponding outputs from all compared methods. The evaluation was conducted on an in-house crowdsourcing platform with 50 participants, yielding a total of 6{,}375 votes, where each scene–prompt pair was evaluated by three distinct annotators. For each scene–prompt pair, annotators were shown the reference scene and all candidate outputs, and were asked to assess each method according to two criteria: (i) text alignment and (ii) scene alignment. The scores in Table~\ref{tab:QuantitiveCompare} indicate the percentage of times each method was selected in the user study. The votes were normalized using a temperature parameter adjustment to ensure fair comparability among different methods.

\begin{table}[t]
    \centering

    \begin{minipage}[t]{0.48\textwidth}
        \centering
        \caption{Ablation study of correspondence-guided attention loss.}
        \label{tab:ABStudyofAttentionLoss}
        \begin{tabular}{@{}lcc@{}}
        \toprule
        Methods & G2.5F-SA$\uparrow$ & G2.5F-TIA$\uparrow$ \\
        \midrule
        w/o CG-Attn Loss     & 9.532 & 8.569 \\
        w/ CG-Attn Loss      & \textbf{9.811} & \textbf{9.639} \\
        \bottomrule
        \end{tabular}
    \end{minipage}%
    \hfill
    \begin{minipage}[t]{0.48\textwidth}
        \centering
        \caption{Ablation study of data construction strategies.}
        \label{tab:comp_data}
        \begin{tabular}{@{}lcc@{}} 
        \toprule
        Data & G2.5F-SA$\uparrow$ & G2.5F-TIA$\uparrow$    \\
        \midrule
        DL3DV-10K~\cite{ling2024dl3dv} & 9.048 & 8.960  \\
        ours w/o filter & 9.648 & 9.079 \\
        ours  & \textbf{9.811} & \textbf{9.639} \\
        \bottomrule
        \end{tabular}
    \end{minipage}
    
\end{table}

\begin{figure}[h]
    \centering
    \begin{minipage}[t]{0.5\textwidth}
        \centering
        \includegraphics[width=\linewidth]{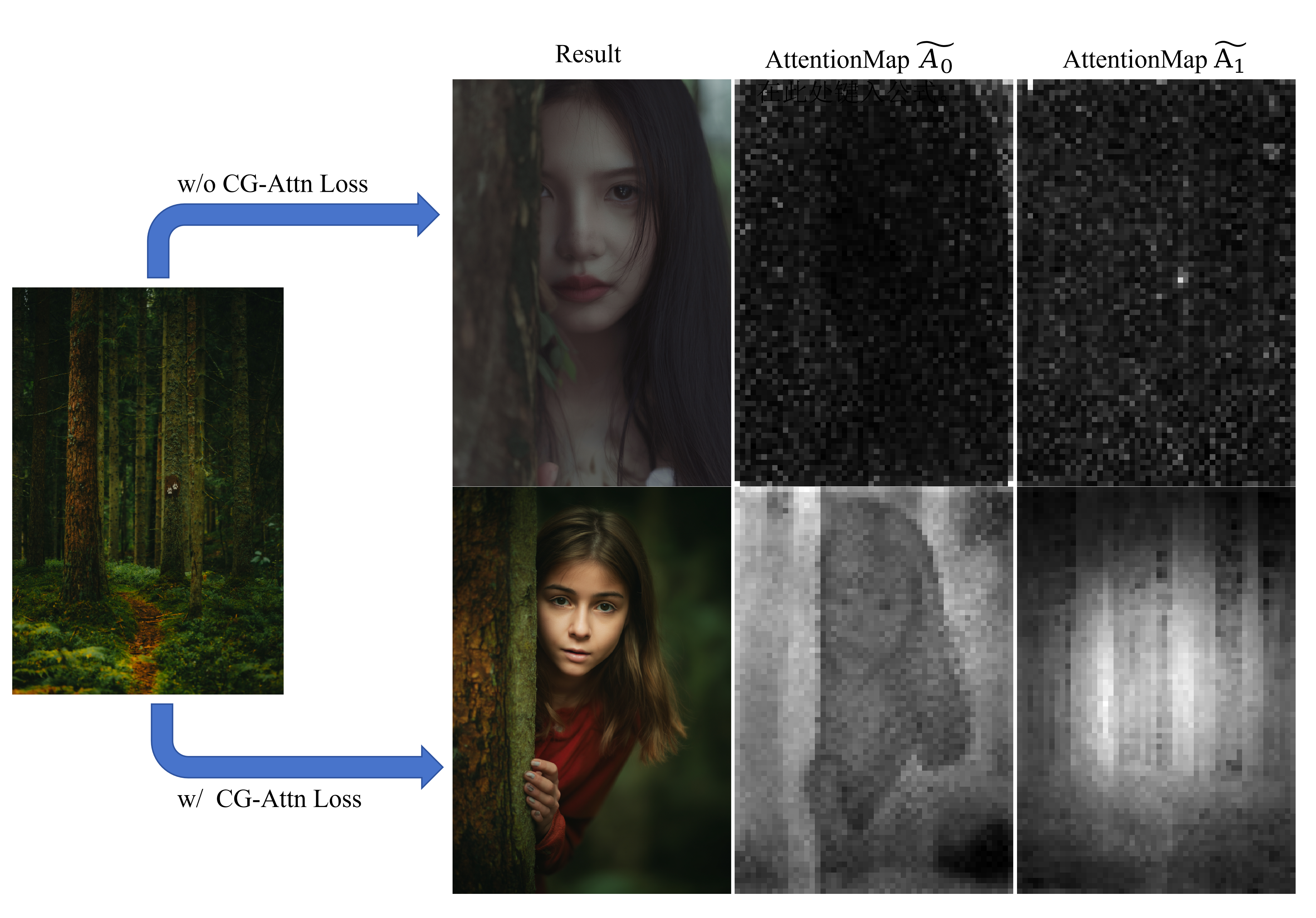}
        \caption{Generated image and attention map comparison with and without correspondence-guided attention loss. (text prompt:a girl hiding behind a tree, close up)}
        \label{fig:attn_map_comparison}
    \end{minipage}
    \hfill
    \begin{minipage}[t]{0.46\textwidth}
        \centering
        \includegraphics[width=\linewidth]{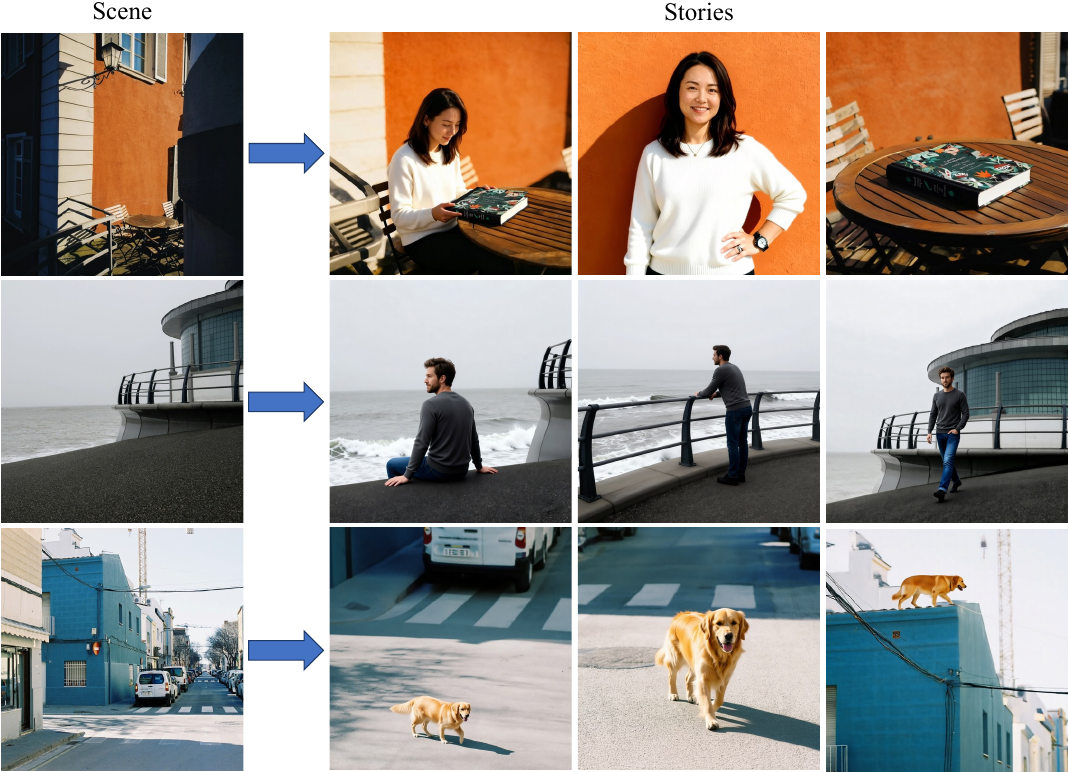}
        \caption{Coherent visual storyboarding. Multi-frame storylines (right) generated from reference scenes (left). Combining scene staging with identity replacement maintains both scene and actor consistency.}
        \label{fig:story}
    \end{minipage}
\end{figure}

As shown in Table~\ref{tab:QuantitiveCompare}, our method achieves the highest preference under both text alignment and scene alignment. Compared to Qwen-Image-Edit~\cite{wu2025qwen}, 
our approach is more frequently judged to produce images that both follow the textual instructions and preserve the underlying scene environment.
We attribute these gains primarily to the combined effect of our scene-consistent data construction pipeline and correspondence-guided attention loss. The former supplies training pairs with  spatially plausible entity–scene relationships and cross-view variations, while the latter explicitly regularizes the spatial layout, thereby validating the effectiveness of our joint design for improving scene consistency.

\subsection{Ablation Study}
In this section, we conduct ablation studies to evaluate the effectiveness of the two proposed components: correspondence-guided attention loss and  scene-consistent data construction pipeline.

\noindent\textbf{Correspondence-Guided Attention Loss.}
Furthermore, the impact of the proposed correspondence-guided attention loss can be observed in Table~\ref{tab:ABStudyofAttentionLoss}, which highlights its role in enhancing the coherence of generated scenes.
With the attention loss, the model achieves improvements on both scene alignment and text--image alignment metrics, with G2.5F-SA increasing by 0.279 and G2.5F-TIA increasing by 1.07, which demonstrates its ability to maintain stable spatial relationships within the scene.
In addition, we provide a visual comparison of generations with and without the proposed correspondence-guided attention loss and analyze the attention maps from the designated mid-level block used to compute the loss, as shown in Figure~\ref{fig:attn_map_comparison}.
$\tilde{A}_0$ denotes the aggregated cross-view attention map for the target image (view 0), while $\tilde{A}_1$ enotes the aggregated cross-view attention map for the scene image (view 1);
brighter regions indicate higher correlation at the corresponding positions.
With the attention loss, the attention maps highlight spatially corresponding regions in both images, indicating that the network has learned to focus on areas crucial for maintaining scene consistency.
These observations collectively demonstrate that the attention loss not only improves visual scene consistency but also guides the model to attend to structurally relevant background regions.

\noindent\textbf{Scene-Consistent Data Construction.}
We compare models trained with our Scene-Consistent Data Construction pipeline (Section~\ref{subsec:data_synthesis}) against a baseline built from real-world multi-view datasets with post-hoc entity insertion. 
For the baseline, we start from multi-view scene sequences in DL3DV-10K~\cite{ling2024dl3dv} and insert foreground humans or objects into the target views using FLUX.1 Kontext Dev~\cite{labs2025flux}, while keeping the network architecture, optimization hyper-parameters, and training schedule identical. 
As reported in Table~\ref{tab:comp_data}, training on our scene-consistent data yields consistently higher G2.5F-SA and G2.5F-TIA scores than the DL3DV-10K–based baseline, demonstrating the clear advantage of our pipeline: while the DL3DV-10K baseline provides multi-view coverage, the post-hoc insertions lack the organic interaction between entity and scene. The resulting compositing artifacts and  instance-scale imbalance likely degrade supervision quality, as further discussed in the supplementary material.

Furthermore, to validate the necessity of our filtering strategy in mitigating structural drift, we ablated training on MINIMA-filtered data against raw, unfiltered data. As shown in Table~\ref{tab:comp_data}, training on unfiltered data contaminated with hallucinations or layout shifts results in a decrease of G2.5F-SA from 9.811 to 9.648 and a more pronounced drop in G2.5F-TIA from 9.639 to 9.079. These results suggest that our filtering step effectively mitigates the impact of generated artifacts, encouraging the model to learn from more consistent samples and thereby suppressing structural drift.

\subsection{Application: Coherent Visual Storyboarding}
As discussed in Section~\ref{sec:Introduction}, continuous visual storytelling necessitates coherence across two primary dimensions: the "stage" and the "actor". While our primary contribution successfully addresses the critical bottleneck of scene consistency, we also explore a practical workaround to achieve comprehensive narrative coherence by incorporating existing identity-preserving techniques.

To demonstrate this, we introduce a cascaded pipeline for comprehensive visual storyboarding. First, we employ our proposed Scene Staging method to synthesize structurally accurate frames, embedding a text-specified entity into the reference scene. Subsequently, we adopt Qwen Image Edit as a post-processing step to replace the generated entity, thereby enforcing strict identity preservation of the "actor".

As illustrated in Figure~\ref{fig:story}, this cascaded approach demonstrates the feasibility of generating multi-frame storylines. The left column shows the reference scenes, while the right columns present the generated storyboards. The "stage" (e.g., the building facades, the seaside, and the street) is robustly preserved by our method, and the specific identities of the "actors" (e.g., the woman, the book, the man, and the dog) are maintained via the subsequent replacement step.

While this decoupled solution serves as a viable strategy to bridge the gap between scene staging and full visual storytelling, it inherently relies on a multi-step process. This observation highlights the necessity for future research into end-to-end frameworks that jointly optimize for both actor and scene consistency.

%% file: sec/5_conclusion.tex
\section{Conclusion}

This paper presents a novel framework explicitly optimized for structural scene preservation, generating images that maintain the underlying spatial layout of a reference scene while synthesizing foreground entities aligned with textual descriptions. Our contributions include: (i) a scene-consistent data construction pipeline that produces high-quality, scene-consistent paired data with precise text–image alignment, enabling effective training with explicit scene consistency constraints; and (ii) a correspondence-guided attention loss that leverages cross-view spatial correspondences to regularize intermediate attention maps, constraining the model to attend to structurally relevant background regions. Extensive experiments across diverse scenarios demonstrate that our method achieves superior performance over existing approaches in both automatic metrics and user studies, with clear improvements in scene alignment and text--image consistency. Qualitative visualizations further confirm that our correspondence-guided design effectively captures fine-grained scene semantics and maintains high scene consistency even under significant viewpoint changes.

Despite these advances, a current limitation of our approach is that it does not explicitly integrate foreground identity preservation into the core model architecture. Achieving comprehensive visual storyboarding currently relies on a decoupled, two-stage pipeline to maintain the "actor". Nevertheless, in principle, the proposed formulation could be extended to jointly model scene and character consistency by augmenting the data construction pipeline and training objectives with additional identity-aware constraints, which we leave as an important direction for future work.

%% file: sec/X_suppl.tex
\section{Evaluation Metrics Selection}
\label{sec:metric_select}

\subsection{Text-Image Alignment Metric Selection}

For evaluating text–image alignment, we compare two metrics: \textbf{CLIP-T}~\cite{gal2022image} and \textbf{Gemini 2.5 Flash Text–Image Alignment (G2.5F-TIA)}~\cite{comanici2025gemini}. CLIP-T is a CLIP-based metric that computes global feature similarity between visual and textual inputs. Leveraging large-scale image–text pairs during training, CLIP-T excels at capturing coarse semantic correspondence between captions and images. However, in our experiments, we identified two notable limitations of CLIP-T in the context of fine-grained generative evaluation:

\begin{enumerate}
    \item \textbf{Limited spatial relationship understanding.} CLIP-T is less sensitive to the relative positioning of objects or scene elements, which can lead to high similarity scores even when spatial arrangements deviate from the textual description.
    \item \textbf{Limited action verification capability.} CLIP-T struggles to assess whether described actions or dynamic attributes in the text are correctly depicted in the image, often failing to capture the precise poses or object interactions that define the action.
\end{enumerate}
To address these limitations, G2.5F-TIA employs a multimodal reasoning approach that integrates spatial and action-aware alignment, yielding evaluations that better reflect fine-grained adherence to the textual prompt.

\begin{figure}[!t]
    \centering
    \begin{minipage}[t]{0.44\textwidth}
        \centering
        \includegraphics[width=\linewidth]{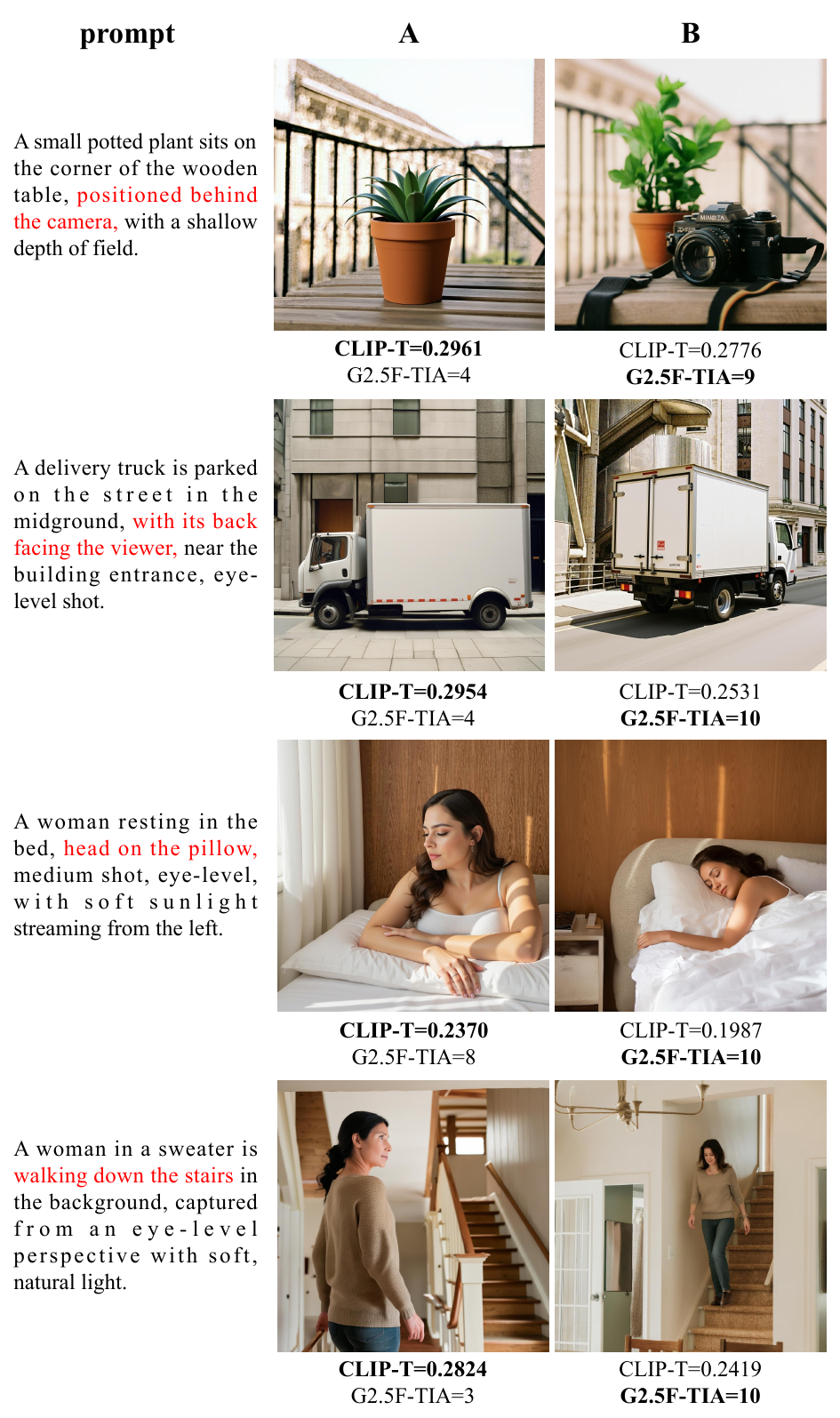}
        \caption{Comparison of text–image alignment metrics (CLIP-T~\cite{gal2022image}$\uparrow$ vs. G2.5F-TIA~\cite{comanici2025gemini}$\uparrow$). Bold scores indicate the preferred set (A or B). Red text highlights spatial (first two) and action (last two) elements. Unlike CLIP-T, G2.5F-TIA effectively penalizes spatial and action violations, better aligning with human perception.
        }
       \label{fig:clipt_compare}
    \end{minipage}
    \hfill
    \begin{minipage}[t]{0.52\textwidth}
        \centering
        \includegraphics[width=\linewidth]{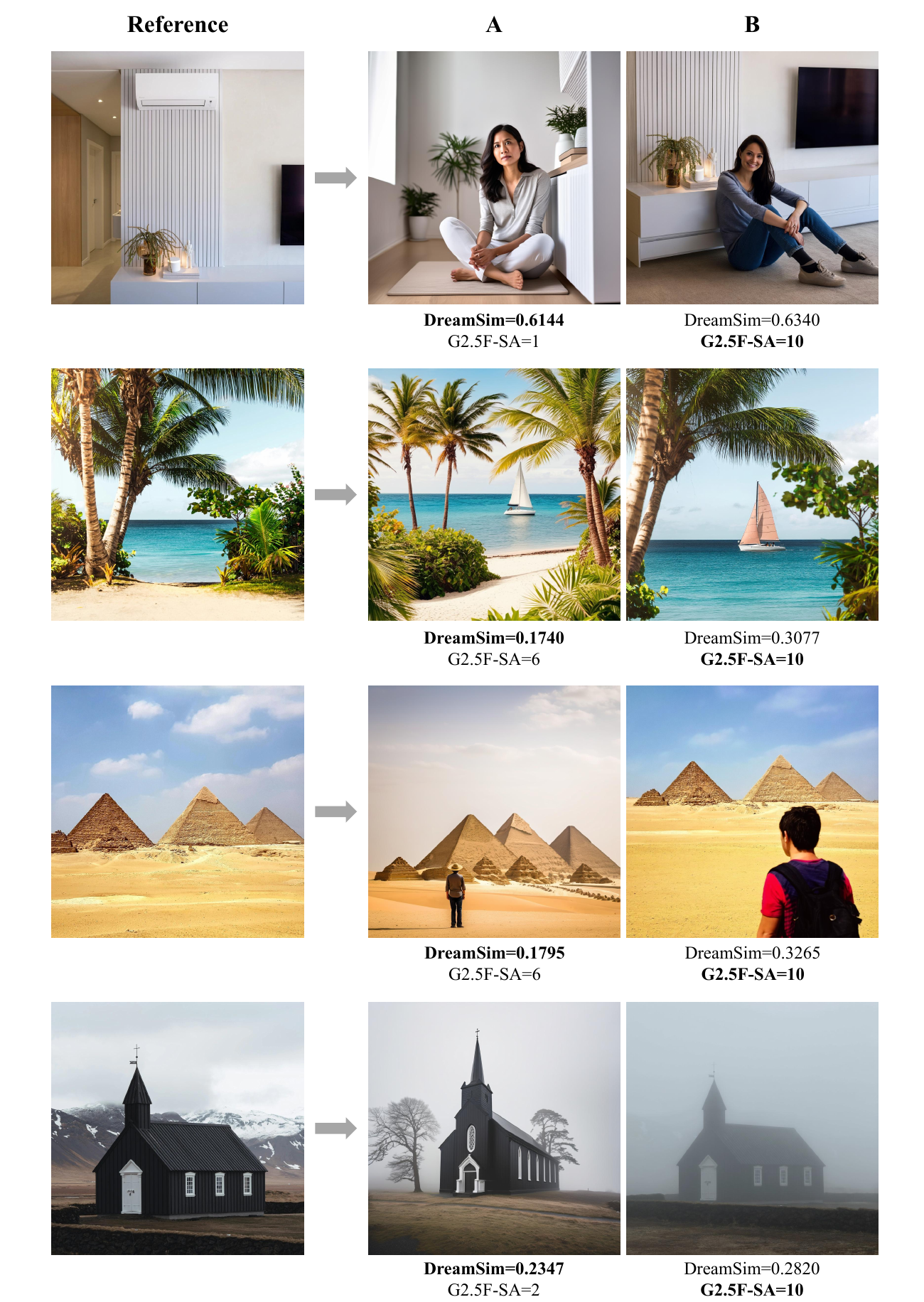}
        \caption{Comparison of scene-alignment metrics (DreamSim~\cite{fu2023dreamsim}$\downarrow$ vs. G2.5F-SA~\cite{comanici2025gemini}$\uparrow$). Two experiment sets (A and B) are shown; in Set B, DreamSim reports higher distances despite the images being visually closer to the reference, whereas G2.5F-SA remains robust to scaling, viewpoint shifts, and defocus blur.}
       \label{fig:dreamsim_compare}
    \end{minipage}
\end{figure}

We present comparative cases, with two sets of generated results (Set A and Set B) shown in Figure~\ref{fig:clipt_compare}, where scores highlighted in bold indicate the set preferred by the respective metric (i.e., higher similarity depending on the convention). In the first two examples, the input text describes explicit spatial composition requirements for objects in the scene, with the spatial composition elements highlighted in red for clarity. Although the image in Set A violates the described spatial arrangement, CLIP-T still produces a high similarity score due to the presence of the correct object categories, whereas G2.5F-TIA assigns a lower score, accurately reflecting the mismatch in spatial configuration. In the latter two cases, the input text specifies an action attribute, with the action-related segments highlighted in red for clarity. The image in Set A fails to depict the described action, yet CLIP-T assigns a high score. G2.5F-TIA, however, correctly penalizes the violation, yielding scores that align more faithfully with human judgment and demonstrating greater sensitivity to fine-grained prompt adherence, particularly in scenarios involving spatial relationships and action verification.

\begin{table}[t]
  \caption{Pairwise accuracy between quantitative evaluation metrics and user-study preferences for different tasks (higher is better).}
  \centering
  \begin{tabular}{@{}lcc@{}}
    \toprule
    Task & Metric & Pairwise Accuracy$\uparrow$ \\
    \midrule
    \multirow{2}{*}{Text-Image Alignment} 
      & G2.5F-TIA & 83.36\% \\
      & CLIP-T & 41.64\% \\
    \midrule
    \multirow{2}{*}{Scene Alignment} 
      & G2.5F-SA & 95.63\% \\
      & DreamSim & 90.85\% \\
    \bottomrule
  \end{tabular}
  \label{tab:metric_pairwise_accuracy}
\end{table}

To statistically validate these observations, we quantified the alignment of each metric with human preference. We compute the proportion of valid comparison pairs (pairs with clear human preference) in which the best image selected by quantitative evaluation metric is the same as the best image selected by participants in the user study. As reported in Table~\ref{tab:metric_pairwise_accuracy}, G2.5F-TIA achieves a pairwise accuracy of \textbf{83.36\%}, significantly outperforming CLIP-T, which only reaches \textbf{41.64\%}. This substantial gap confirms that CLIP-T often misaligns with human judgment on fine-grained details, whereas G2.5F-TIA serves as a robust proxy for human evaluation.

\subsection{Scene Alignment Metric Selection}
To evaluate scene alignment in our experiments, we considered several perceptual similarity metrics, including \textbf{DreamSim}~\cite{fu2023dreamsim} which is adopted in SceneDecorator~\cite{song2025scenedecorator} for measuring scene alignment and the multimodal large model \textbf{Gemini 2.5 Flash}~\cite{comanici2025gemini}. DreamSim achieves holistic similarity assessment by jointly modeling foreground and background features. While effective for general similarity tasks, its architecture exhibits particular sensitivity to foreground variations. This characteristic becomes problematic for dynamic scene composition tasks involving frequent object-level modifications, where foreground alterations may disproportionately influence the similarity score regardless of background preservation. The Gemini 2.5 Flash Scene Alignment (G2.5F-SA) metric addresses this architectural constraint through explicit feature decoupling, providing more balanced evaluation of scene edits while maintaining robust performance on conventional similarity.

Qualitative comparisons with DreamSim and G2.5F-SA are provided in Figure~\ref{fig:dreamsim_compare}. We perform scene-consistent image generation from the reference image and produce two sets of results, denoted as Set A and Set B. For each set, the similarity scores computed by DreamSim and G2.5F-SA between the reference and the generated images are displayed below the corresponding samples. Scores that are highlighted in bold indicate the set preferred by the respective metric (higher similarity depending on the metric's convention). Notably, DreamSim considers the images generated in Set B less similar to the reference, despite their visual closeness. As illustrated, common generative transformations including explicit object insertions as well as camera zoom operation, or background blur, can lead to such discrepancies in DreamSim’s evaluation. In contrast, G2.5F-SA explicitly accounts for these permissible variations, yielding evaluations that align more faithfully with human perceptual judgments.

We further corroborate this selection through quantitative correlation with user preferences, employing the same pairwise accuracy calculation protocol as described in the text--image alignment section. As summarized in Table~\ref{tab:metric_pairwise_accuracy}, G2.5F-SA achieves a pairwise accuracy of \textbf{95.63\%} with human judgments, surpassing DreamSim's \textbf{90.85\%}. While DreamSim remains a highly competitive baseline, its holistic metric couples both foreground and background elements, making it slightly less suited for our task where foreground modifications (e.g., inserting new entities) are explicitly allowed while background consistency must be preserved. G2.5F-SA bridges this gap by focusing specifically on background preservation, yielding evaluations that more faithfully reflect human perception of scene consistency.

\section{Extended Details on Evaluation Metrics and User Study}
This section provides additional details on the automatic metrics and user study protocol used in the main paper. Specifically, we expand upon the evaluation setup introduced in Section 4.2 (main paper) and the user study discussed in Section 4.3 (main paper). We explicitly describe the Gemini 2.5 Flash prompts used for automatic scoring and the annotation interface shown to human raters.

\begin{table*}[h]
    \centering
    \small
    \renewcommand{\arraystretch}{1.3}
    \begin{tabular}{p{0.15\linewidth} | p{0.8\linewidth}}
        \toprule
        \textbf{Metric} & \textbf{Input Prompt Template} \\
        \midrule
        \textbf{G2.5F-SA} \newline (Scene Alignment) & 
        From scale 0 to 10: \newline
        Give one score for how well the second image remains the same physical scene as the first image. \newline
        Do not penalize normal edits: object additions/removals, close-up / background blur (DOF), zoom/rotation, and moderate viewpoint changes. \newline
        (0 = completely different scene; 10 = clearly the same scene with only allowed edits.) \newline
        \newline
        Put the score in a list such that output score = [score]. \newline
        \newline
        Text Prompt (use text only to understand intended edits): \{prompt\}
        \\
        \midrule
        \textbf{G2.5F-TIA} \newline (Text-Image Alignment) & 
        From scale 0 to 10: \newline
        A score from 0 to 10 will be given based on the success in following the prompt. \newline
        (0 indicates that the image frames do not follow the prompt at all. 10 indicates the image frames follows the prompt perfectly.) \newline
        \newline
        Put the score in a list such that output score = [score]. \newline
        \newline
        Text Prompt: \{prompt\}
        \\
        \bottomrule
    \end{tabular}
    \caption{\textbf{Prompts for Automatic Evaluation.} We employ Gemini 2.5 Flash to evaluate scene alignment and text-image alignment. The placeholders \{prompt\} are replaced with the specific textual instruction for each test case.}
    \label{tab:gemini_prompts}
\end{table*}

\subsection{Details of Automatic Metrics}
\label{sec:supp_auto_metrics}

As described in Section 4.2 (main paper), we utilize Gemini 2.5 Flash~\cite{comanici2025gemini} as an evaluator to assess both scene alignment (G2.5F-SA) and text-image alignment (G2.5F-TIA). The exact prompts used to query the VLM are detailed in Table~\ref{tab:gemini_prompts}. 

For scene alignment, the prompt explicitly instructs the model to tolerate acceptable modifications such as object modifications and viewpoint changes, focusing the score on the preservation of the underlying scene. For text-image alignment, the prompt focuses solely on semantic adherence to the text instruction.

\subsection{Details of User Study}
\label{sec:supp_user_study}

To ensure a rigorous subjective evaluation, we conducted a user study involving 50 participants on an in-house crowdsourcing platform. The interface was designed to facilitate direct comparison between methods while maintaining a focus on the specific criteria defined in our task.

\paragraph{Interface Layout}
For each evaluation trial, the interface presented the reference image on the far left, with the target text instruction displayed at the top. To the right of the reference image, four candidate results generated by the compared methods were displayed in a horizontal row. The order of the candidate methods was randomized for every trial to prevent positional bias.

\paragraph{Annotation Instructions}
Participants were asked to evaluate the results based on a "Select All" mechanism. For each of the two criteria below, annotators were instructed to select all candidate images (Indices 1--4) that satisfied the requirement. The specific guidelines provided to the annotators were as follows:

\begin{enumerate}
    \item \textbf{Text Alignment:} 
    \begin{quote}
    \textit{Select all images from candidates 1--4 that are consistent with the textual description.}
    \end{quote}
    
    \item \textbf{Scene Alignment:} 
    \begin{quote}
    \textit{Select all images from candidates 1--4 that remain consistent with the reference scene.}
    \par\vspace{0.4em}
    \noindent\small{\textbf{Guideline:} Changes in viewpoint and composition are allowed. As long as the structure and details of the scene (e.g., positions of walls/furniture, decorative elements) remain unchanged, the image should be considered scene-consistent.}
    \end{quote}
\end{enumerate}

\paragraph{Data Aggregation Strategy}
To ensure evaluation reliability, each sample was assessed by three annotators, and we filtered the results to retain only methods with majority agreement. The final preference percentages in Table 1 of the main paper were then derived using the temperature-scaled normalization scheme described therein. Specifically, for a prompt where $k$ methods are retained after consensus filtering, we assign a score contribution $w = 1/k^{\alpha}$ (where $\alpha=0.8$ serves as the temperature factor) to balance between approval and fractional voting, followed by normalization.

\section{Visual Analysis of DL3DV-10K–based Composited Images}

\begin{figure}[t]
  \centering
  \includegraphics[width=0.7\linewidth]{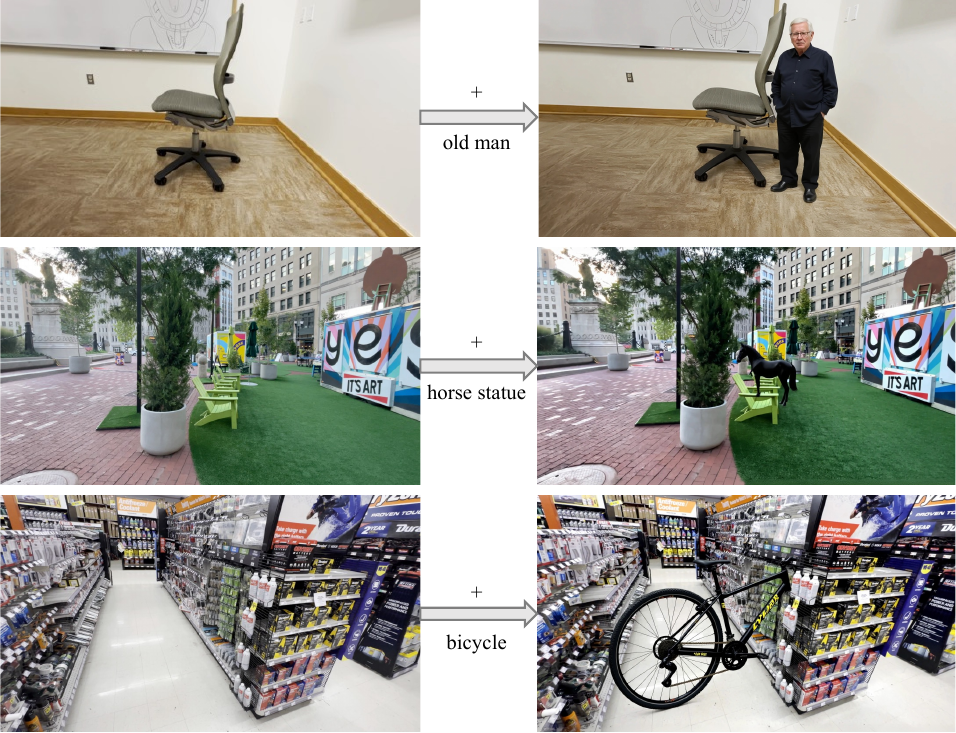}
  \caption{Examples of the DL3DV-10K–based data-construction baseline.
For each example, the left image is the original DL3DV-10K scene image, and the right image is the composited result after inserting a new entity into the scene.
  }
  \label{fig:dl3dv_qual}
\end{figure}

We visualize some examples of the DL3DV-10K–based data construction baseline in Figure~\ref{fig:dl3dv_qual}, where entities (an old man, a horse statue, and a bicycle) are composited into real scenes.
Across these cases, the inserted instances are frequently mis-scaled relative to nearby objects: in the first row, the person is rendered at approximately the same height as the adjacent chair, which is clearly implausible; in the second row, the horse statue is slightly under-scaled and fails to rest on the floor, instead seeming to hover around the chairs with incorrect spatial relationships; and in the third row, the bicycle is erroneously inserted so that its structure intersects the shelving unit.
In addition, imperfect matting and lighting inconsistencies around object boundaries introduce visible halos and shading artifacts that break the photometric coherence of the scene. These instance-scale imbalances and compositing artifacts confirm that, although DL3DV-10K provides genuine multi-view coverage, the resulting images offer noisy supervision signals for scene-consistent image generation, consistent with the lower G2.5F-SA and G2.5F-TIA scores reported in Table 3 of the main paper compared to our scene-consistent data construction pipeline. 

\begin{figure*}[!t]
  \centering
  \includegraphics[width=1\textwidth]{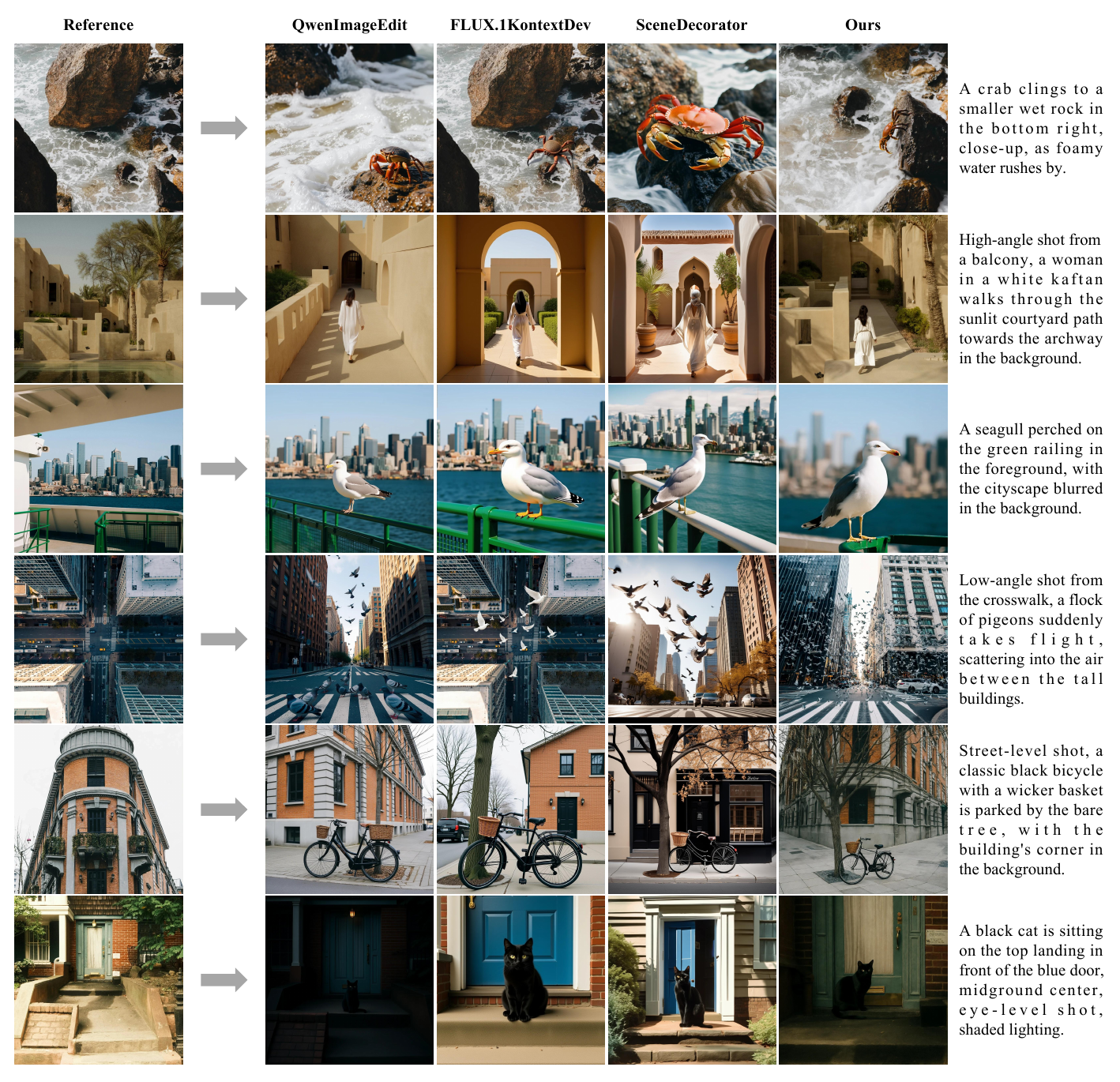}
  \caption{Additional qualitative comparisons of our method with Qwen-Image-Edit~\cite{wu2025qwen}, FLUX.1 Kontext Dev~\cite{labs2025flux}, and SceneDecorator~\cite{song2025scenedecorator}. Our method better preserves spatial alignment and fine-grained scene semantics while avoiding unintended background changes.}
  \label{fig:qualitative-comparison-supp}
\end{figure*}
\section{Additional Qualitative Comparison Results}
\label{sec:qualitative_result_suppl}

We provide additional qualitative comparisons with Qwen-Image-Edit~\cite{wu2025qwen}, FLUX.1 Kontext Dev~\cite{labs2025flux}, and SceneDecorator~\cite{song2025scenedecorator} in Figure~\ref{fig:qualitative-comparison-supp}. These examples cover diverse scene staging scenarios and further highlight the limitations of the compared approaches, including unintended background changes, reduced spatial accuracy, and appearance inconsistencies in scene elements. Our method consistently preserves fine-grained scene semantics and spatial alignment, producing outputs that remain faithful to the source scene while adhering closely to the textual instructions.

\section{Supplementary Videos}

To further demonstrate the effectiveness of our framework in maintaining scene consistency across diverse viewpoints, and to illustrate its application in video storyboarding, we provide two videos in our \href{https://drive.google.com/drive/folders/11W4XEzquyM0Jx3gdfPrIVamX_xk6x4yb?usp=sharing}{online supplementary folder}. These videos were synthesized using the Kling image-to-video model, utilizing keyframes produced by our method. In the first sequence, the start frame is generated by our method while the final frame is the original scene. Conversely, the second sequence begins with the original scene, with the intermediate and final frames being generated views. Across both videos, the interpolated trajectories—including long-range camera motions such as zoom-outs and large viewpoint changes—remain stable. Crucially, the scene structure is preserved without noticeable texture flickering or structural drift, highlighting the suitability of our method as a robust keyframe generator for long-duration video synthesis.